\documentclass[11pt,a4paper]{article}
\usepackage[hyperref]{acl2020}
\usepackage{times}
\usepackage{latexsym}
\usepackage{url}
\usepackage{proof}
\usepackage{xspace}
\usepackage{multirow}
\usepackage{tabu} %rowfont
\usepackage{CJKutf8} 
\usepackage{balance}
\usepackage{forest}
\usepackage{caption}
\usepackage{subcaption}
\usepackage{rotating}
\usepackage{dsfont}
\usepackage{pifont}
\usepackage{bm} 
\usepackage{amsmath}
\usepackage{amssymb}
\usepackage{amsthm}
\usepackage{algorithm}
\usepackage{array}
\usepackage{float}
\usepackage{stfloats}
\usepackage{hhline}
\usepackage[noend]{algpseudocode}
\definecolor{dollarbill}{rgb}{0.52, 0.73, 0.4}
\definecolor{deepmagenta}{rgb}{0.8, 0.0, 0.8}
\definecolor{coralred}{rgb}{1.0, 0.25, 0.25}
\definecolor{chromeyellow}{rgb}{1.0, 0.65, 0.0}
\definecolor{ao(english)}{rgb}{0.0, 0.5, 0.0}
\definecolor{gray}{rgb}{0.5, 0.5, 0.5}
\definecolor{mediumturquoise}{rgb}{0.28, 0.8, 0.9}

\newcommand{\ytop}{y_{\text{top}}\xspace}
\newcommand{\ptop}{p_{\text{top}}\xspace}

\newcommand{\notes}[1]{}%{\it {\small {#1}}}}

\theoremstyle{definition}

\theoremstyle{plain}

\newcommand{\ith}[1]{\ensuremath{i^{{th}}}}

\newcount\permx
\newcount\permy
\def\permdot#1#2{
\permx=#1 \advance\permx by-1
\permy=#2 \advance\permy by-1
\psframe[fillcolor=black, fillstyle=solid]
(\permx,\permy)(#1, #2)
}

\newcommand{\boxnum}[1]{{\setlength{\fboxsep}{1pt}\raisebox{1pt}{\hspace{1pt}\fbox{\tiny #1}\hspace{1pt}}}}
\newcommand{\ind}[1]{\ensuremath{_{\kern-0.5pt\boxnum{#1}}}}

\newcommand{\vecx}{\ensuremath{\bm{x}}\xspace}
\newcommand{\vecy}{\ensuremath{\bm{y}}\xspace}

\newcommand{\smallnt}[1]{\ensuremath{_{\mbox{\tiny PP}}}\xspace}

\newcommand{\pseudocode}{Algorithm}
\floatname{algorithm}{\pseudocode}

\iffalse

\else

\fi

\newcommand{\eos}{\mbox{\scriptsize \texttt{</s>}}\xspace}
\newcommand{\sos}{\mbox{\scriptsize \texttt{<s>}}\xspace}

\newcommand{\READ}{{$R\!E\!A\!D()$}}
\newcommand{\rot}[1]{\rotatebox{90}{\hspace{-0.1cm}{#1}}}
\newcolumntype{C}[1]{>{\centering\let\newline\\\arraybackslash\hspace{0pt}}m{#1}}
\newcommand\NoThen{\renewcommand\algorithmicthen{}}
\usepackage{microtype}

\aclfinalcopy % Uncomment this line for the final submission

\newif\iflong

\longtrue
\title{
    Simultaneous Translation Policies: From Fixed to Adaptive 
    }

\author{Baigong Zheng $^{1}$ \, 
        Kaibo Liu $^{1}$ \,
        Renjie Zheng $^{1}$ \,
        Mingbo Ma $^{1}$ \, \\
        {\bf
        Hairong Liu $^{1}$ \,
        Liang Huang $^{1, 2}$ } \\
  $^{1}$Baidu Research, Sunnyvale, CA, USA \\
  $^{2}$Oregon State University, Corvallis, OR, USA \\
  {\small \texttt{ \{baigongzheng, kaiboliu, renjiezheng, mingboma\}@baidu.com} } \\ 
  {\small \texttt{ \{liuhairong, lianghuang\}@baidu.com} } \\
  }

\date{}

\begin{document}
\maketitle
\begin{abstract}
Adaptive policies are better than fixed policies for simultaneous translation, 
since they can flexibly balance 
the tradeoff between translation quality and latency
%reading source tokens against writing target tokens 
based on the current context information.
%Simultaneous translation prefers adaptive policies rather than fixed policies, for the latter could either make aggressive predictions or cause high latency during translation process.
But previous methods on obtaining adaptive policies either rely on complicated training process, or underperform simple fixed policies. 
We design an algorithm to achieve adaptive policies via a simple heuristic composition of a set of fixed policies. 
Experiments on Chinese$\to$English and German$\to$English show that our adaptive policies can outperform fixed ones by up to 4 BLEU points for the same latency, and more surprisingly, it even surpasses the BLEU score of full-sentence translation in the greedy mode (and very close to beam mode), but with much lower latency.
\end{abstract}

\section{Introduction}
% !TEX root = main.tex

Simultaneous translation (ST) aims to provide good translation quality while keeping the latency of translation process as low as possible. This is very important for the scenarios that require simultaneity, such as international summits and negotiations. For this, human interpreters usually start translation before the source sentence ends. 
However, this makes the translation process much more challenging than the full-sentence translation, because to balance the translation quality and latency, interpreters need to make decisions on when to continue translation and when to stop temporarily to wait for more source side information, which are difficult, especially for syntactically divergent language pairs, such as German and English.

\begin{figure}[t]
\centering
\includegraphics[width=0.9\linewidth]{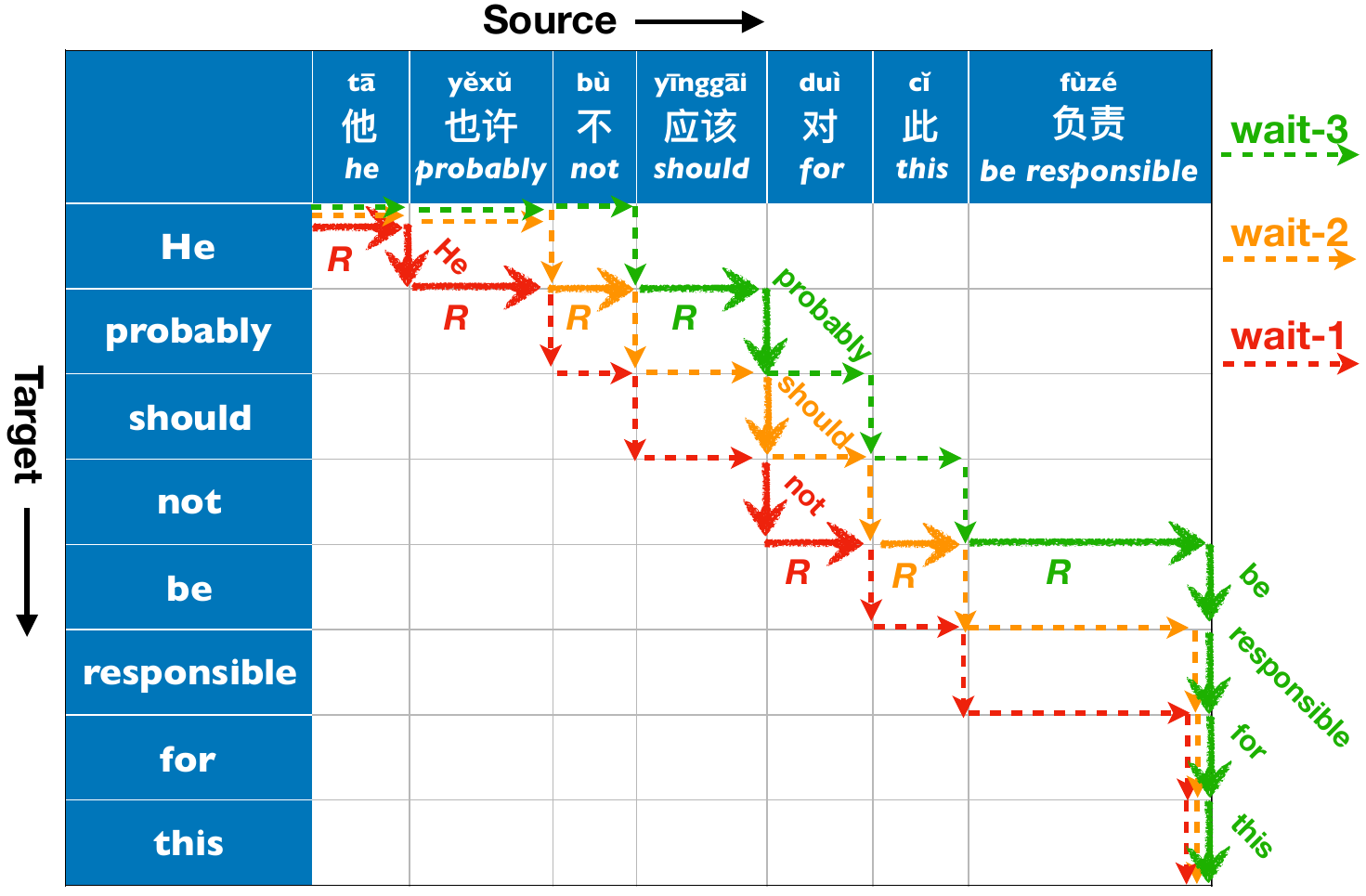}
\caption{An adaptive policy (in bold arrows) composed of three wait-$k$ policies ($k =  1, 2, 3$). }
\label{fig:compstn}
\vspace{-10pt}
\end{figure}

The above decisions can be considered as two actions: READ (wait for a new source word) and WRITE (emit a translated target word)~\cite{gu+:2017}. Then we only need to decide which action to choose at each step, and the solution can be represented by a {\em policy}.
Earlier works~\cite{mahsa+:2013, bangalore+:2012, fugen+:2007, sridhar+:2013, jaitly2016online} study policies as a part of speech-to-speech ST system, where the policies usually try to separate the source sentence into several chunks that can be translated safely.
Recent works focus on obtaining policies for text-to-text ST, which can be generally divided into two categories: fixed and adaptive.
Fixed policies~\cite{ma+:2019, dalvi+:2018} usually follow some simple rules to choose actions. For example, the wait-$k$ policy by~\citet{ma+:2019} first chooses $k$ READ actions, and then chooses WRITE and READ alternatively.
This kind of policies do not utilize the context information and can be either too aggressive or too conservative in different cases. 

By contrast, adaptive policies try to make decisions on the fly using the currently available information. It is obvious that this kind of policies is more desirable for ST than the fixed ones, and  different methods are explored to achieve an adaptive policy. 
The majority of such methods~\cite{grissom+:2014, Cho+:16, gu+:2017, alinejad+:2018, zheng+:2019b} are based on full-sentence translation models, which may be simple to use but cannot outperform fixed policies applied with ``genuinely simultaneous" models trained for ST~\cite{ma+:2019}. Other methods~\cite{arivazhagan+:2019, zheng+:2019a} try  to learn a policy together with the underlying translation model, but they rely on complicated and time-consuming training process.

In this paper, we propose to achieve an adaptive policy via a much simpler heuristic composition of a set of wait-$k$ policies (e.g., $k\!=\!1 \!\sim\! 10$). 
See Fig.~\ref{fig:compstn} for an example. To further improve the translation quality of our method, we apply ensemble of models trained with different wait-$k$ policies. 
Our experiments on Chinese$\to$English and German$\to$English translation 
show that our method can achieve up to 4 BLEU points improvement over the wait-$k$ method for same latency. 
More interestingly, compared with full-sentence translation, our method achieves higher BLEU scores than greedy search but with much lower latency, and is close to the results from beam search.

\section{Preliminaries}

% TEX root = main.tex

%full-sentence NMT consists of an encoder and a decoder
\paragraph{Full-sentence translation.}
Neural machine translation (NMT) model usually consists of two components: an encoder, which encodes the source sentence $\vecx=(x_1, \dots, x_m)$ into a sequence of hidden states, and a decoder, which sequentially predicts target tokens conditioned on those hidden states and previous predictions. The probability of the predicted target sequence $\vecy=(y_1,\dots,y_n)$ will be 
%\vspace{-5pt}
\[ 
%\vspace{-5pt}
p(\vecy \mid \vecx) = \textstyle\prod_{t=1}^{|\vecy|}  p(y_t \mid \vecx,\, \vecy_{<t})
\]
where $\vecy_{<t}=(y_1,\dots, y_{t-1})$ denotes the target sequence predicted before step $t$.

%prefix-to-prefix training 
%However, simultaneous translation is different from the full-sentence translation, for we are required to keep low latency of the translation process, implying that we usually need to make predictions conditioned on partial source sentence instead of the whole sentence. 
%Previous works~\cite{Cho+:16, gu+:2017, zheng+:2019b} apply full-sentence translation model to do simultaneous translation, resulting in a mismatch between training and testing.
\paragraph{Simultaneous translation.}
\citet{ma+:2019} propose a {\em prefix-to-prefix} framework to train models to make predictions conditioned on partial source sentences.
%, alleviating the mismatch between training and testing. 
In this way, the probability of predicted sequence $\vecy$ becomes
%\vspace{-5pt}
\[
%\vspace{-5pt}
p_g(\vecy \mid \vecx) = \textstyle\prod_{t=1}^{|\vecy|} p(y_t \mid \vecx_{\leq{g(t)}},\, \vecy_{<t})
\]
where $g(t)$ is a monotonic non-decreasing function of $t$, denoting the number of processed source tokens when predicting $y_t$.  
%wait-$k$ method training and inference policy
%One important component in a model for simultaneous translation is its {\em policy}, represented by $g(t)$ in the above equation, which decides at each time step whether to read a new source token or to emit the current predicted target token. 
This function $g(t)$ can be used to represent a policy for ST.
%For example, the full-sentence translation can be represented by $g(t) = |\vecx|$.
\citet{ma+:2019} introduce a kind of fixed policies, called wait-$k$ policy, that can be defined by the following %function 
%One example is the wait-$k$ policy introduced by~\citet{ma+:2019}, which first waits $k$ source tokens and then outputs predicted tokens concurrently with the rest of source sentence. This can be defined by 
%\vspace{-5pt}
\[
%\vspace{-5pt}
g_k(t) = \min\{ |\vecx|, t+k-1\}.
\]
Intuitively, this policy first waits $k$ source tokens and then outputs predicted tokens concurrently with the rest of source sentence. 
%encoder catchup policy
%For language pairs where the length difference of source and target sentences is large, \citet{ma+:2019} propose to catchup the longer sentence during translation process. 
%For example, if source sentence is longer, then with catchup parameter $c=4$ we will read one additional source token every 4 prediction steps. 

%\section{Algorithm}
% !TEX root = main.tex

\section{Obtaining an Adaptive Policy}\label{sec:method}

\begin{figure}[t]
\centering
\includegraphics[width=.9\linewidth]{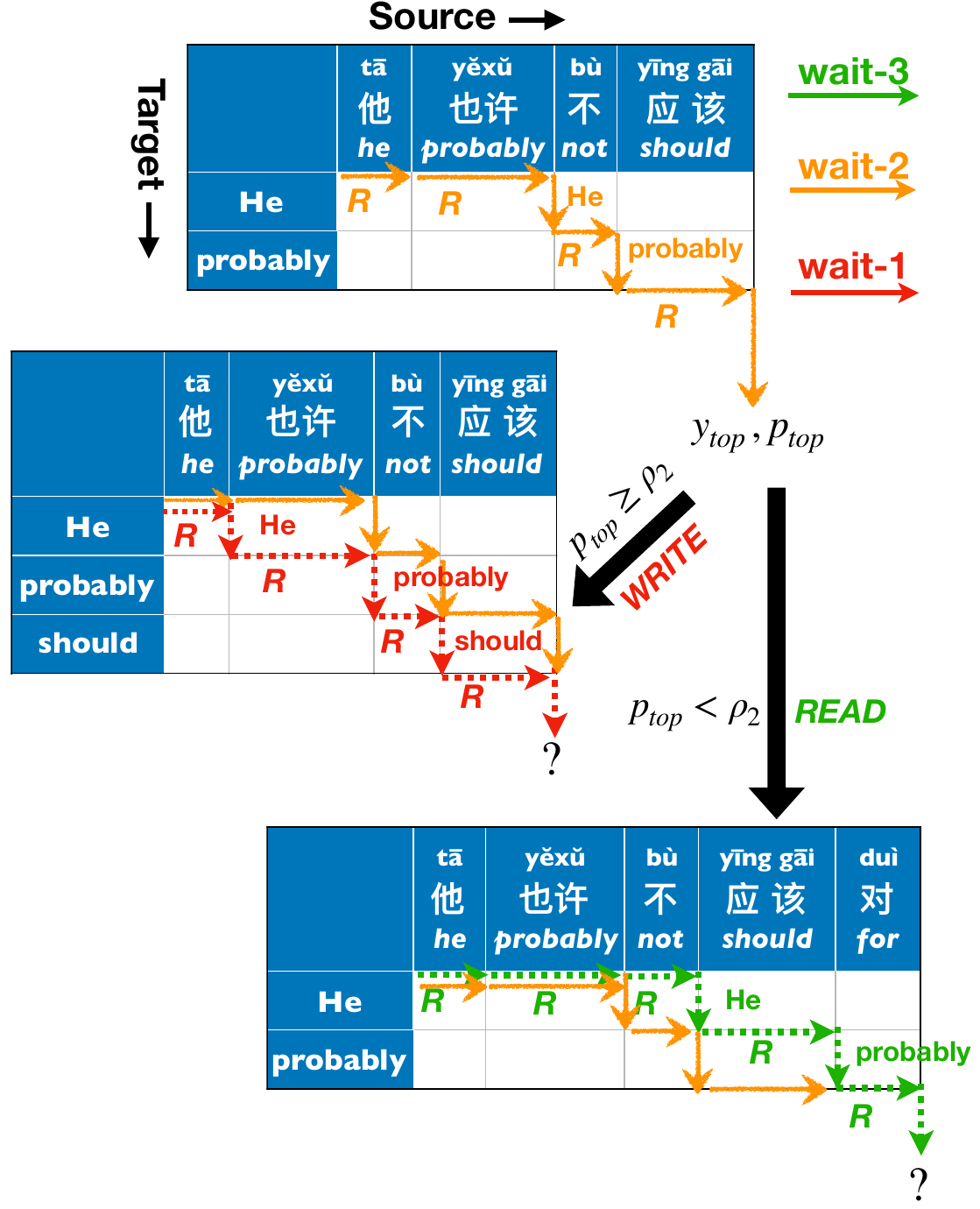}
\vspace{-5pt}
\caption{Choose actions based on model confidence. 
%A new policy is applied after choosing an action. 
In this example, we will choose an action based on the top probability $\ptop$, and apply a new policy (the dotted arrows) after the chosen action.
%apply either wait-1 or wait-3 policy based on the token distribution $D$ obtained from wait-2 policy.
}
\label{fig:example}
\vspace{-10pt}
\end{figure}
Assume we have a set of wait-$k$ policies and the corresponding models $M_k$ ($k = k_{\min} \dots k_{\max}$). 
We can obtain an adaptive policy, whose lag at each step is between $k_{\min}$ and $k_{\max}$, meaning that at each step, the target sequence falls behind the source sequence at most $k_{\max}$ tokens and at least $k_{\min}$ tokens.
At each step, there is a wait-$k$ policy synchronizing the adaptive policy, meaning that they have the same lag at that step.  
%Assume we have an adaptive policy, whose lag at each step is between $k_{\min}$ and $k_{\max}$, meaning that at each step, the target sequence falls behind the source sequence at most $k_{\max}$ tokens and at least $k_{\min}$ tokens.
%Then at each step, there is a wait-$k$ policy synchronizing the adaptive policy, meaning that they have the same lag at that step. Therefore, we could use a set of wait-$k$ policies to simulate the adaptive policy for integer $k_{\min} \le k \le k_{\max}$. 
Specifically, at any step $t$, if the lag of the adaptive policy is $k'$, then we apply the NMT model %to follow 
with the wait-$k'$ policy and force it to predict existing target tokens until step $t$, when the model will make a new prediction as the output of step $t$.

However, 
%the set of wait-$k$ policies does not provide decisions on READ and WRITE actions.  In another words, 
the above method only shows how to simulate the adaptive policy to make a prediction at one step if we would like to write at that step, but it does not tell us at which steps we should write. 
%To solve this problem, 
We utilize the model confidence to make such a decision. 
Specifically, we set a probability threshold $\rho_k$ for each wait-$k$ policy. At each step, if the NMT model follows a wait-$k'$ policy, and predicts the most likely token with probability higher than the threshold $\rho_{k'}$, then we consider the model is confident on this prediction, and choose WRITE action; otherwise, we choose READ action.  
Figure~\ref{fig:example} gives an example for this process.

We define the process of applying a wait-$k$ model $M_k$ with a wait-$k$ policy on a given sequence pair $(\vecx, \vecy)$ 
by the following
%\vspace{-5pt}
\[
%\vspace{-5pt}
\ytop, \, \ptop \leftarrow P_k(M_k, \vecx, \vecy)
\]
which forces model $M_k$ to predict $\vecy$, and returns the top token $\ytop$ at the final step with the corresponding probability $\ptop$. 
The process of reading and returning a new source token is denoted by \READ, and
expression $\vecx \circ x$ represents to append an element $x$ to the end of sequence $\vecx$.
We denote by \sos and \eos the start symbol and end symbol of a sequence. Then Algorithm~\ref{alg:adaptive} gives the pseudocode of the above method. % applied with model $M$.  
\begin{algorithm}
\caption{ST decoding with an adaptive policy}
\label{alg:adaptive}
\begin{algorithmic}
\State {\bf Input}:  
two integers $k_{\min}$ and $k_{\max}$,
a set of 
\\ \indent \indent \ \ \ \ 
NMT models $M_k$, 
and a sequence of 
\\ \indent \indent \ \ \ \
thresholds $\rho_k$ for $k_{\min} \le k \le k_{\max}$.
%\While {$y_{|\vecy|} \ne \eos$}
%    \If {$x_{|\vecx|} = \eos$} 
%    \State $k \leftarrow k_{\max}$
%    \Else
%    \State $k \leftarrow |\vecx|-|\vecy|$
%    \EndIf
%    \If {$x_{|\vecx|} \ne \eos$ and $k < k_{\min} $} 
%    \State $Read(\vecx)$
%    \Else
%        \State $D \leftarrow P_k(M, \vecx, \vecy)$
%        \If {$x_{|\vecx|} \ne \eos$ and $k < k_{\max}$ and \indent \indent $\max(D) < \rho_k$}
%        \State $Read(\vecx)$
%        \Else
%        \State $Write(\vecy, \argmax(D))$
%        \EndIf
%    \EndIf
%\EndWhile
\While {$x_{|\vecx|} \ne \eos$ and $y_{|\vecy|} \ne \eos$}
    \State $k \leftarrow |\vecx|-|\vecy|$
    \State $\ytop, \, \ptop \leftarrow P_k(M_k, \vecx, \vecy)$
    \NoThen
    \If {$k \ge k_{\max}$ or ($k \ge k_{\min}$ and $\ptop \ge \rho_k$)}
        \State $\vecy \leftarrow \vecy \ \circ \ \ytop$ \Comment{Write action}
    \Else
        \State $\vecx \leftarrow \vecx \  \circ\ $\READ \Comment{Read action}
    \EndIf
\EndWhile
\While {$y_{|\vecy|} \ne \eos$}
    \State $\ytop, \, \ptop \leftarrow P_{k_{\max}}(M_{k_{\max}}, \vecx, \vecy)$
    \State $\vecy \leftarrow \vecy \ \circ \ \ytop$ \Comment{Write action}
\EndWhile
\Return $\vecy$
\end{algorithmic}
\end{algorithm}
%\vspace{-5pt}

\section{Ensemble of Wait-$k$ Models}
Using the corresponding model $M_k$ with each wait-$k$ policies may not give us the best performance. %, and it is natural to choose the best model trained in the prefix-to-prefix framework for each policy.  
If we have a set of models trained independently with different wait-$k$ policies, then we can apply ensemble of those models~\cite{dietterich:2000, hansen+:1990} to improve the translation quality, which is also used to improve the translation quality of full-sentence translation~\cite{stahlberg+:2017}.
However, there may be two issues to apply ensemble of all models: (1) the runtime for each prediction could be longer, resulting in higher latency; and (2) the translation accuracy may be worse, for the best model for one policy may give bad performance when doing inference with another policy. 
To avoid these, we propose to apply ensemble of the top-3 models for each policy. That is, we first generate distribution with the top-3 models independently with the same policy, and then take the arithmetic average of the three distributions as the final token distribution at that step.

\section{Experiments}\label{sec:exp}
% !TEX root = main.tex

\begin{figure*}[t]
\centering
\includegraphics[width=.45\linewidth]{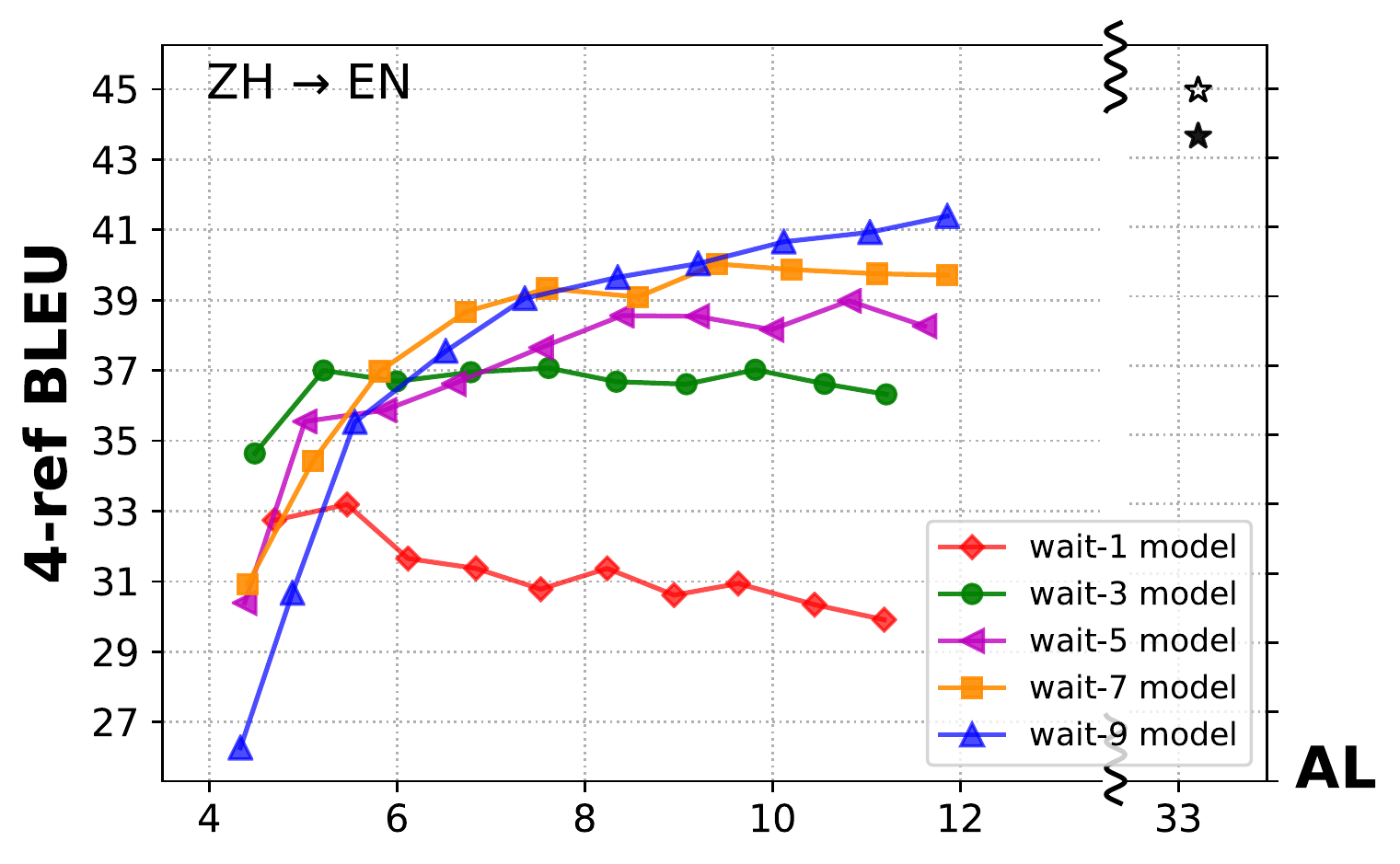} 
\includegraphics[width=.45\linewidth]{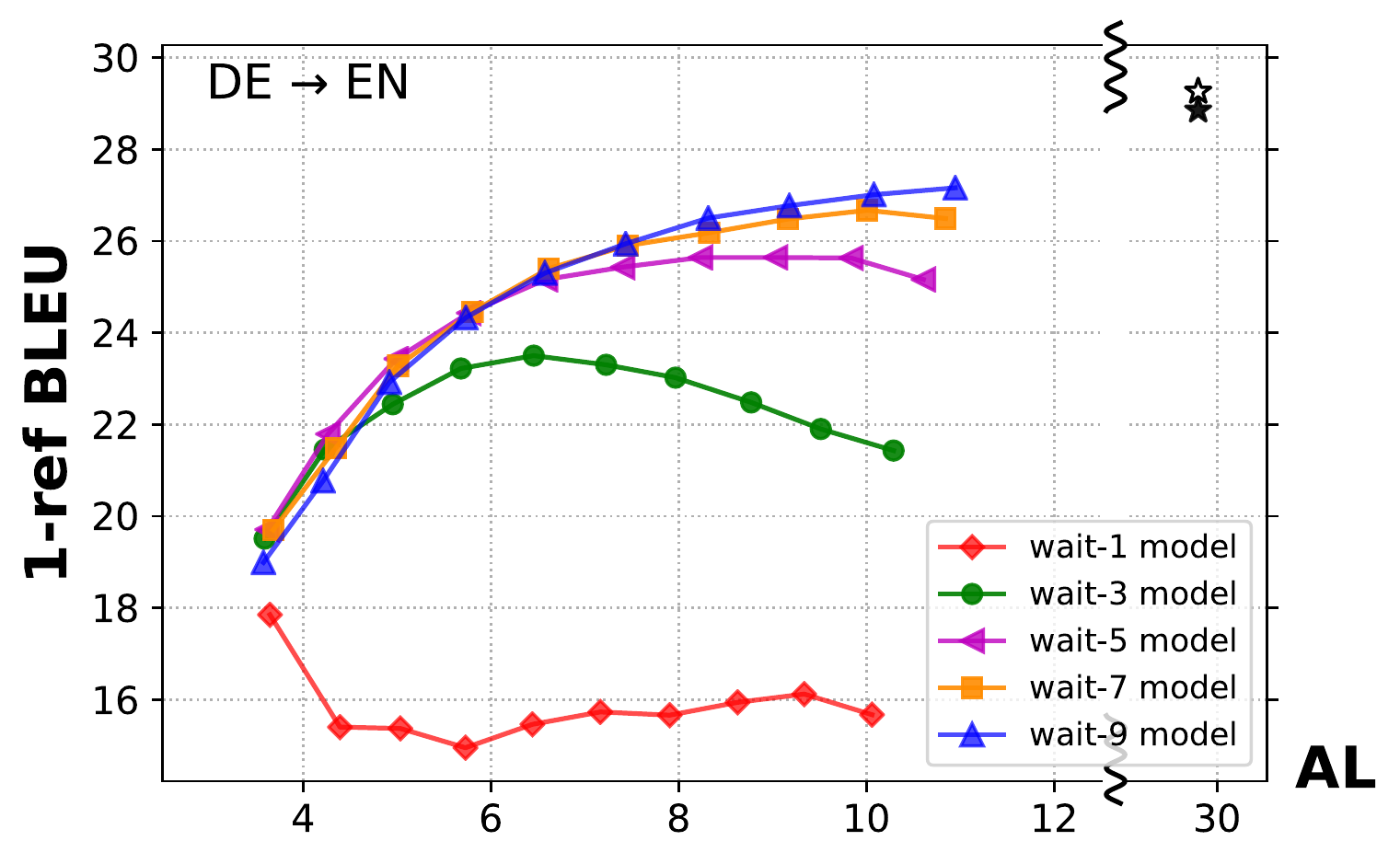}
%\vspace{-10pt}
\caption{
    Performance of models with different policies on dev set.
    Each model is trained with one wait-$k$ policy (i.e. wait-$k$ model) and tested with ten different wait-$k'$ policies for integer $1\le k' \le 10$.
    Each line corresponds to one model.
    %\textcolor{coralred}{$\mdblkdiamond$}: wait-1 model, 
    %\textcolor{ao(english)}{$\mdsmblkcircle$}: wait-3 model,
    %\textcolor{deepmagenta}{$\smallblacktriangleleft$}: wait-5 model,
    %\textcolor{chromeyellow}{$\mdsmblksquare$}: wait-7 model,
    %\textcolor{blue}{$\blacktriangle$}: wait-9 model,
    %\textcolor{black}{$\bigstar$} $\bigwhitestar$
    \ding{72} \ding{73}: full-sentence translation with greedy search and beam search (beam size = 10) respectively. 
    %\textcolor{black}{$\bigstar$}: full-sentence translation with greedy search,
    %$\bigwhitestar$: full-sentence translation with beam search (beam size = 10).
}
%\vspace{-5pt}
\label{fig:policy}
\end{figure*}

\paragraph{Datasets and models.}
%We conduct experiments on two language pairs: Chinese$\to$English (ZH$\to$EN) and German$\to$English (DE$\to$EN). 
We conduct experiments on Chinese$\to$English (ZH$\to$EN) and German$\to$English (DE$\to$EN) translation. 
For ZH$\to$EN, we use NIST corpus (2M sentence pairs) as training set, NIST 2006 as dev set, and NIST 2008 as test set. 
For DE$\to$EN, we use WMT15 parallel corpus %\footnote{\url{http://www.statmt.org/wmt15/translation-task.html}} 
for training, newstest-2013 for validation and newstest-2015 for testing. 
All datasets are tokenized and segmented into sub-word units with byte-pair  encoding~\cite{sennrich+:2016}.  
We take Transformer-base~\cite{vaswani+:2017} as our model architecture, and follow~\citet{ma+:2019} to train our model with wait-$k$ policies for integer $1\le k \le 10$.
In the following experiments, we only use catchup~\cite{ma+:2019} for DE$\to$EN translation, where we read one additional source token after every 6 predictions. 
We use BLEU~\cite{BLEU:2002} as the translation quality metric, and Average Lagging (AL)~\cite{ma+:2019} as the latency metric, which measures the lag behind source in terms of the number of source tokens.

\paragraph{Performance with different policies.}

%In their work, 
%\citet{ma+:2019} train an NMT model with wait-$k$ policy and test it with the same policy. 
%However, we are curious about 
We first evaluate the performance of each model with different policies, 
%i.e. those wait-$k'$ policies for $k' \ne k$, 
which helps us to choose models for different policies. 
Specifically, we apply each model with ten different wait-$k$ policies on dev set to compare the performance. Fig.~\ref{fig:policy} shows the results of five models. % trained with $k \in \{1, 3, 5, 7, 9 \}$.
We find the best model for one policy may not be the one trained with that policy. For example, on ZH$\to$EN translation, 
the best model for wait-1 policy is the one trained with wait-3 policy.
Further, there is no one model could achieve the best performance for all policies.

\paragraph{Comparing different methods.}

\begin{figure*}[t]
\centering
\includegraphics[width=.45\linewidth]{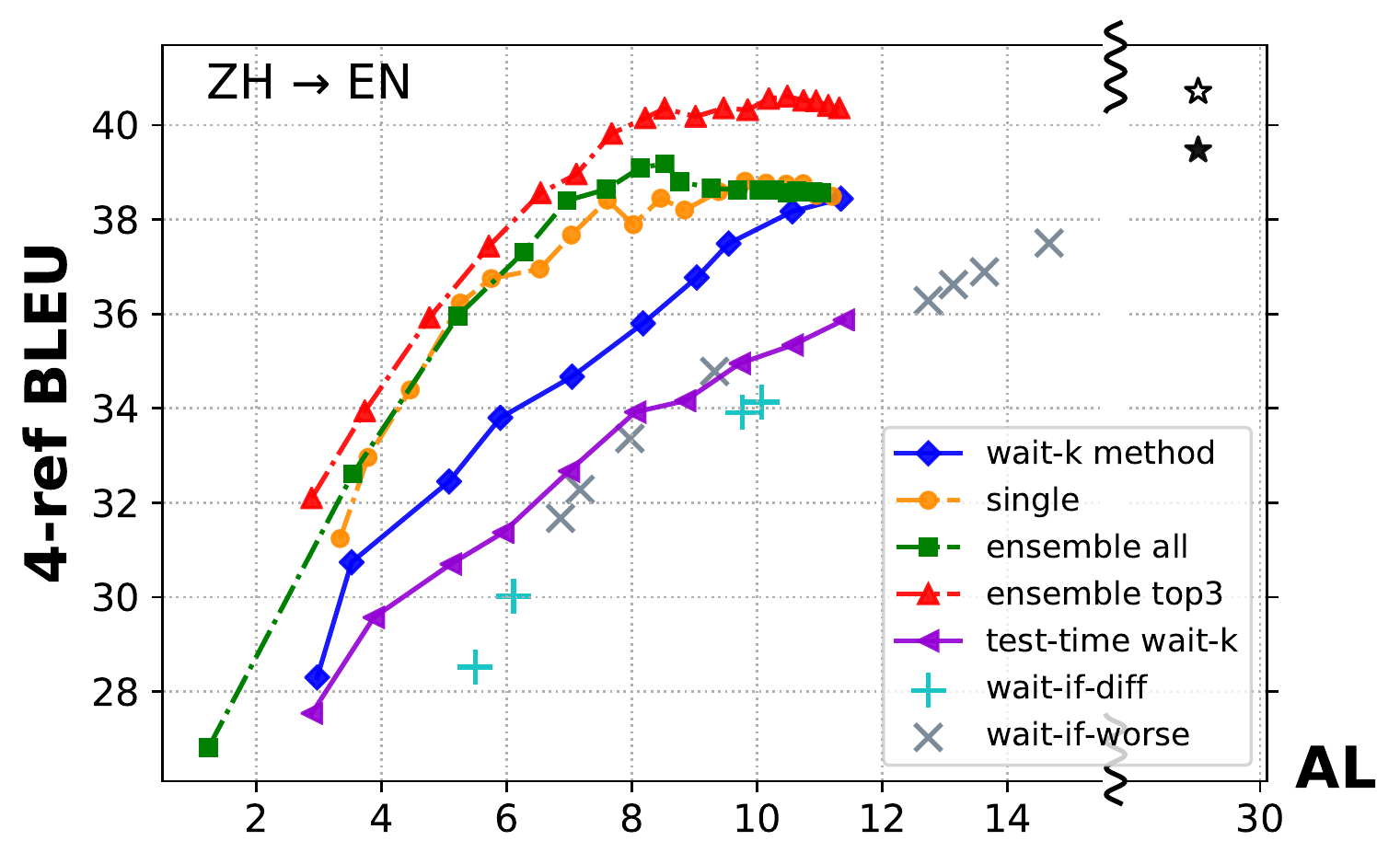}
\includegraphics[width=.45\linewidth]{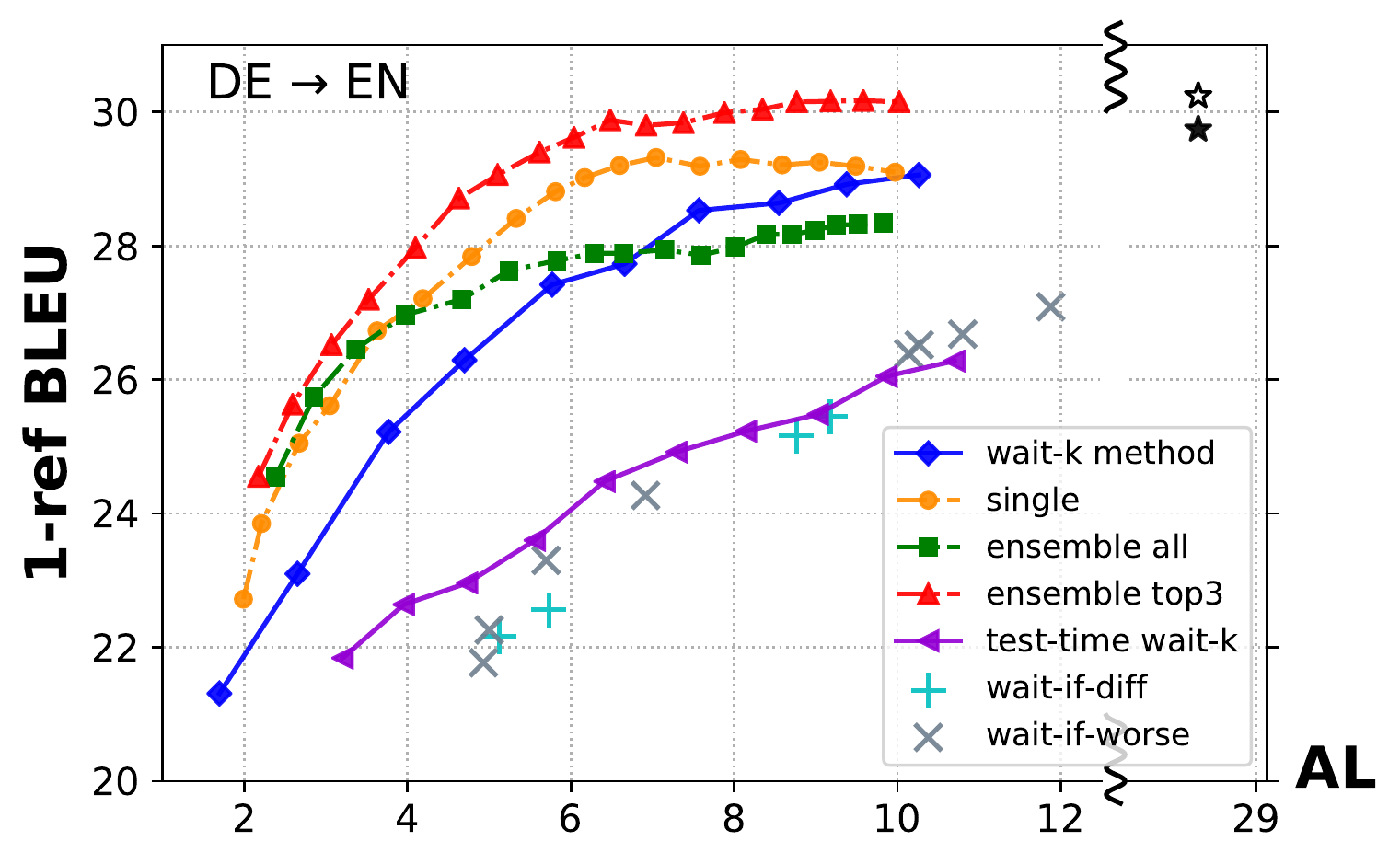}
%\vspace{-5pt}
\caption{Performance of different methods on test set. 
    %Our method achieves higher BLEU scores with same latency and surpass full-sentence greedy search with AL < 10.
    Our {\bf single} method achieves better BLEU scores than {\bf wait-$k$} method with same latency.
    And our {\bf ensemble top-3} method achieves the highest BLEU scores with same latency, and outperforms full-sentence greedy search with AL $<$ 9. 
    %\textcolor{blue}{$\mdblkdiamond$}: wait-$k$ method, 
    %\textcolor{coralred}{$\blacktriangle$}: ensemble top-3,
    %\textcolor{ao(english)}{$\mdsmblksquare$}: ensemble all,
    %\textcolor{chromeyellow}{$\mdsmblkcircle$}: single,
    %\textcolor{deepmagenta}{$\smallblacktriangleleft$}: test-time wait-$k$,
    %\textcolor{mediumturquoise}{$+$}: wait-if-diff for $s_0 \in \{4, 6\}$ and $\delta \in \{2, 4\}$,
    %\textcolor{gray}{$\times$}: wait-if-worse for $s_0 \in \{1, 2, 4, 6\}$ and $\delta \in \{ 1, 2 \}$.
    %\textcolor{black}{$\bigstar$} $\bigwhitestar$
    \ding{72} \ding{73}: full-sentence translation with greedy search and beam search (beam size = 10) respectively. 
    %$\bigwhitestar$: full-sentence translation with beam search (beam size = 10).
}
%\vspace{-15pt}
\label{fig:method}
\end{figure*}

\begin{CJK*}{UTF8}{gbsn}
\begin{table*}[t]
\resizebox{1\linewidth}{!}{%
\setlength{\tabcolsep}{-2pt}
\centering
\begin{tabu}{ C{3cm} | c c c c c c c c c c c c c c c c C{1.5cm} c c c C{1.5cm} c C{2cm} c c c c c c c |  C{4.3cm}  } 
\hline 
\rowfont{\it\small}
\rule{0pt}{12pt}
pinyin \ \ & & w\v{o}m\'en & & \ \ xi\`ang &  &  sh\`ouh\`aizh\v{e} \ & &  de & & \kern-0.5em ji\={a}sh\v{u} \kern-0.5em  & & bi\v{a}osh\`{\i} & & zu\`{\i} & &  \kern-0.5em ch\'engzh\`{\i} \kern-1.3em & & \kern-0.4em de  & & t\'{o}ngq\'{\i}ng \kern-1.4em  & & \kern-0.5em h\'e & &\kern-0.4em  \=ai&  & \kern-1em d\`ao  \kern-1em   & & & & & \\
\rule{0pt}{11pt}
input \ \ &\ \  {\bf ``} \ \ & 我  们 & &  向 &  & 受 害 者  & & 的 & & \kern-0.5em  家 属\kern-0.5em  & & 表 示 & & 最 & & \kern-0.8em 诚 挚 \kern-1.3em  & & \kern-0.4em 的 & & 同 情 \kern-1.4em &  & \kern-0.5em 和 & & \kern-0.4em 哀- & &\kern-1em 悼 \kern-1em   & & \ \ {\bf .}  & & {\bf "}  \ \ & \\
gloss \ \ & & we  & &  to &  & victim  & & 's & & \kern-0.5em  family\kern-0.5em  & & express & & most & & \kern-0.5em sincere \kern-1.3em  & & \kern-0.4em 's & & sympathy \kern-1.4em &  & \kern-0.5em  and & & \multicolumn{3}{c}{\kern-0.5em condolence}   & & & &    & \\
\hline
\rule{0pt}{11pt}
ensemble top-3 $\rho_1\!=\!1, \rho_{10}\!=\!0$ (AL=7) &  & & &  &  & & {\bf ``} & & we &  & & & & & & & & & &  & express our &  & most sincere sympathy and & & &  & &  & &  &  \ \ condol- ences to the families of the victims {\bf . "}\\
\hline
\rule{0pt}{11pt}
ensemble top-3 $\rho_1\!=\!0.4, \rho_{10}\!=\!0$ (AL=2.8) &  & & {\bf ``}&   & we &   & &  & &   & &  & &  & &  &  express the most sincere  & & &   & sympathy to the  &  & families of the &    & &  & victims {\bf .} & & &   &  {\bf "}\\
\hline
wait-3 (AL=3.72)  &   & & &  & {\bf ``}  & & \kern-0.5em we  & & \ have \kern-0.5em &  & \kern-0.2em offered \kern-1em  & & our  & & best \kern-0.5em  & & wishes & & \ \ to & & the  & & families & & \ \ of &  & the &   & victims &  & {\bf , "} he said {\bf .}\\ 
\hline 
\multirow{1}{*}{\shortstack{full-sentence \\ translation \\ (AL=16) }} &   & & &  &  & & & & &  & & & & & & & & & & & & & & & & & &  &  & & {\bf ``} we express the most sincere sympathy and condol- ences to the families of the victims {\bf . "} \\
 \hline
\end{tabu} 
}
\caption{
    One example from ZH$\to$EN dev set. 
    %AL of ensemble top-3 ($\rho_1=1, \rho_{10}=0$) is 7, AL of ensemble top-3 ($\rho_1=0.4, \rho_{10}=0$) is 2.8, AL of wait-3 method is 3.72, and AL of full-sentence translation is 16. 
    Although {\bf wait-3} method has low latency, it makes anticipations on ``offered" and ``wishes", and adds additional words ``he said", which are not accurate translation. Our {\bf ensemble top-3} method could provide better translation with lower latency.
}
\label{tab:exp}
\end{table*} 
\end{CJK*}

\begin{table}[h]
\resizebox{0.99\linewidth}{!}{%
\centering
\begin{tabular} {c | c@{\ }c | c@{\ }c}
\multirow{2}{*}{Method} & \multicolumn{2}{c | }{ZH$\to$EN} & \multicolumn{2}{c}{DE$\to$EN} \\
& BLEU  & AL & BLEU & AL \\
\hline
Full-sentence (greedy) & 39.47 & 29.551   & 29.74 & 28.581 \\
Full-sentence (beam)   & 40.71 & 29.551   & 30.24 & 28.581 \\
Ensemble Top-3                & 40.15 & 8.209    & 30.15 & 8.766
\end{tabular}
}
\caption{Compare our method with full-sentence translation. Our {\bf ensemble top-3} method could outperform the greedy search and get close to beam search (beam size = 10) with lower latency.}
\label{tab:cmp_full}
\end{table}

We compare our method with others from literature: % on translation quality and latency: 
{\bf wait-$k$} method~\cite{ma+:2019} (train and test models with the same wait-$k$ policy), 
{\bf test-time wait-$k$} method~\cite{ma+:2019} (apply full-sentence model with wait-$k$ policies), 
{\bf wait-if-diff}~\cite{Cho+:16} (start with $s_0$ source tokens, choose to read only if top token at $t$-th step diffs from that at $(t-\delta)$-th step), 
and {\bf wait-if-worse}~\cite{Cho+:16} (start with $s_0$ source tokens, choose to read only if the top probability at $t$-th step is smaller than that at $(t-\delta)$-th step). 
For {\bf wait-if-diff} we set $s_0 \in \{4, 6\}$ and $\delta \in \{2, 4\}$; and for {\bf wait-if-worse} we set $s_0 \in \{1, 2, 4, 6\}$ and $\delta \in \{ 1, 2 \}$.

For our method, we test three different cases: (1) {\bf single}, where for each policy we apply the corresponding model that trained with the same policy; (2) {\bf ensemble top-3}, where for each policy we apply the ensemble of 3 models that achieve the highest BLEU scores with that policy on dev set; (3) {\bf ensemble all}, where we apply the ensemble of all 10 models for each policy.  
For thresholds, we first choose $\rho_{1}$ and $\rho_{10}$, and the other thresholds are computed in the following way: 
$
\rho_i = \rho_{1} - d\cdot(i-1)
$
for integer $1 \le i \le 10$ and $d=(\rho_{1} - \rho_{10}) / 9$.
We test with $\rho_1 \in \{$0.2, 0.3, 0.4, 0.5, 0.6, 0.7, 0.8, 0.9$\}$, $\rho_{10}=0$ and $\rho_1 = 1$, $\rho_{10} \in \{$0, 0.1, 0.2, 0.3, 0.4, 0.5, 0.6, 0.7, 0.8, 0.9$\}$, totally 18 different settings in our experiments. 
The reason behind these settings is that we assume our adaptive policy cannot be either too aggressive or too conservative (as mentioned at the beginning of Section~\ref{sec:method}). 
The policy is the most aggressive for $k = 1$, so we set $\rho_1$ as the largest; while for $k = 10$ the policy is the most conservative, so we set $\rho_{10}$ the smallest.

The comparison is provided in Fig.~\ref{fig:method} (the corresponding numeric scores are provided in Appendix~\ref{sec:apdx}).
%We can see that {\bf single} is already better than {\bf wait-$k$} method, with the improvement up to 2 BLEU point at some points. And our {\bf ensemble top-3} works best, achieving improvement up to 4 BLEU points compared with {\bf wait-$k$} method. 
Compared with {\bf wait-$k$} method, 
our {\bf single} method achieves improvement of up to 2 BLEU point, and our {\bf ensemble top-3} achieves improvement up to 4 BLEU points.
Compared with full-sentence translation, our {\bf ensemble top-3} 
surprisingly outperforms greedy search with much lower latency (AL $<$ 9),
and achieves BLEU scores close to that from beam search (see Table~\ref{tab:cmp_full}).
We also give one ZH$\to$EN translation example from dev set in Table~\ref{tab:exp} to compare different methods, % in the appendix, 
showing that our method achieves an adaptive policy with low latency and good translation quality. %and gives the similar result to full-sentence translation.

\paragraph{Efficiency.}
% this speed is faster than real-time speech input, and we argue that the system could be further speedup by parallelization and other more efficient implementation tools or language like C++.  
%Efficiency is crucial for ST system, and 
To evaluate the efficiency, 
we present in Table~\ref{tab:runtime} the averaged time needed to predict one token for different methods. 
These methods are tested on one GeForce GTX TITAN-X GPU
for ZH$\to$EN test set. %Results are shown in Table~\ref{tab:runtime}.
We can see that our {\bf ensemble top-3} method needs about 0.2 seconds to make a prediction on average. %, which is slower than the others. 
However, if the source sentence is revealed in the same speed as general speech, which is about 0.6 seconds per token in Chinese~\cite{zheng+:2019c}, then our method is still faster than that (which means that it could be used for real-time). 
Further, we believe the efficiency of 
our method could be improved with other techniques, such as parallelizing the running of three models in the ensemble, making it less an issue.

%\vspace{-5pt}
% Running time table
\begin{table}[h]
\resizebox{0.98\linewidth}{!}{%
\centering
\begin{tabular} {c | c}
\hline
Method & Time per Token \\
\hline
Full-sentence  & 0.0122 s \\
Wait-3 & 0.0162 s \\
%Wait-9 & 0.0175 s \\
Single ($\rho_1=0.4, \rho_{10}=0$) & 0.1057 s \\
Ensemble Top-3 ($\rho_1=0.4, \rho_{10}=0$) & 0.2085 s \\
%Ensemble Top-3 ($\rho_1=0.8, \rho_{10}=0$) & 0.2080 s \\
%Ensemble Top-3 ($\rho_1=1.0, \rho_{10}=0.5$) & 0.1469 s \\
\hline
\end{tabular}
}
%\vspace{-5pt}
\caption{Averaged time needed by different methods to predict one token on ZH$\to$EN test set.}
%\vspace{-15pt}
\label{tab:runtime}
\end{table}
%\paragraph{Example} 
%We give one translation example from ZH$\to$EN dev set in Table~\ref{tab:exp}, which shows that our method achieves an adaptive policy and gives the similar result to full-sentence translation.

\section{Conclusions}
We have designed a simple heuristic algorithm to obtain an adaptive policy based on a set of wait-$k$ policies, and applied ensemble in our method to improve the translation quality while maintaining low latency. Experiments show that our method not only outperforms the original wait-$k$ method with relatively large gap, but also %achieves higher BLEU scores than greedy full-sentence translation with much lower latency. %One potential issue of our method is its efficiency, and we leave this as future work.
surpasses greedy full-sentence translation with much lower latency.

\section*{Acknowledgments}
We thank the anonymous reviewers for helpful suggestions.

%\newpage
\balance
\bibliography{main}
\bibliographystyle{acl_natbib}

\clearpage
\nobalance
\appendix

\section{Appendices}~\label{sec:apdx}
%\label{sec:appendix}

% !TEX root = main.tex
%\section{Supplemental Material}
%We give one translation example from ZH$\to$EN dev set in Table~\ref{tab:exp}, which shows that our method achieves an adaptive policy and gives the similar result to full-sentence translation.
We provide the complete results of Figure~\ref{fig:method} from Section~\ref{sec:exp} in the following tables, where AL is Average Lagging. Note that for ZH$\to$EN, we use 4-reference BLEU; while for DE$\to$EN we use single-reference BLEU.

\begin{table}[h]
\resizebox{0.99\linewidth}{!}{%
\centering
\begin{tabular} {c@{\ }c | c c | c c}
& \multirow{2}{*}{Hyper-parameters} & \multicolumn{2}{c | }{ZH$\to$EN} & \multicolumn{2}{c}{DE$\to$EN} \\
& & BLEU  & AL & BLEU & AL \\
\hline
\multirow{4}{*}{\rot{wait-if-diff}} & $s_0 = 4, \ \delta = 2$  & 28.52 & 5.493   & 22.16 & 5.121 \\
& $s_0 = 6, \ \delta = 2$  & 30.02 & 6.108   & 22.56 & 5.731 \\
& $s_0 = 4, \ \delta = 4$  & 33.91 & 9.764   & 25.16 & 8.763 \\
& $s_0 = 6, \ \delta = 4$  & 34.13 & 10.075  & 25.45 & 9.177 \\
%\end{tabular}
%}
%\vspace{-5pt}
%\caption{Complete results of {\bf wait-if-diff} method.}%, backing the pluses in Figure~\ref{fig:method}}
%\label{tab:wid}
%\end{table}
%\vspace{-15pt}
%
%% Done
%\begin{table}[h]
%\resizebox{0.9\linewidth}{!}{%
%\centering
%\begin{tabular} {c | c c | c c}
%\multirow{2}{*}{Hyper-parameters} & \multicolumn{2}{c | }{ZH$\to$EN} & \multicolumn{2}{c}{DE$\to$EN} \\
%& BLEU  & AL & BLEU & AL \\
\hline
\multirow{18}{*}{\rot{ensemble top-3}}& $\rho_1 = 0.2, \ \rho_{10} = 0.0$  & 32.10 & 2.880   & 24.55 & 2.171  \\
& $\rho_1 = 0.3, \ \rho_{10} = 0.0$  & 33.94 & 3.729   & 25.63 & 2.592  \\
& $\rho_1 = 0.4, \ \rho_{10} = 0.0$  & 35.92 & 4.762   & 26.52 & 3.068  \\
& $\rho_1 = 0.5, \ \rho_{10} = 0.0$  & 37.43 & 5.710   & 27.20 & 3.523  \\
& $\rho_1 = 0.6, \ \rho_{10} = 0.0$  & 38.56 & 6.538   & 27.97 & 4.096  \\
& $\rho_1 = 0.7, \ \rho_{10} = 0.0$  & 38.96 & 7.109   & 28.71 & 4.628  \\
& $\rho_1 = 0.8, \ \rho_{10} = 0.0$  & 39.82 & 7.675   & 29.06 & 5.101  \\
& $\rho_1 = 0.9, \ \rho_{10} = 0.0$  & 40.15 & 8.209   & 29.40 & 5.616  \\
& $\rho_1 = 1.0, \ \rho_{10} = 0.0$  & 40.35 & 8.520   & 29.62 & 6.038  \\
& $\rho_1 = 1.0, \ \rho_{10} = 0.1$  & 40.18 & 9.013   & 29.88 & 6.482  \\
& $\rho_1 = 1.0, \ \rho_{10} = 0.2$  & 40.36 & 9.462   & 29.80 & 6.923  \\
& $\rho_1 = 1.0, \ \rho_{10} = 0.3$  & 40.32 & 9.848   & 29.84 & 7.379  \\
& $\rho_1 = 1.0, \ \rho_{10} = 0.4$  & 40.56 & 10.185  & 29.99 & 7.882  \\
& $\rho_1 = 1.0, \ \rho_{10} = 0.5$  & 40.61 & 10.480  & 30.04 & 8.347  \\
& $\rho_1 = 1.0, \ \rho_{10} = 0.6$  & 40.52 & 10.739  & 30.15 & 8.766  \\
& $\rho_1 = 1.0, \ \rho_{10} = 0.7$  & 40.51 & 10.939  & 30.16 & 9.182  \\
& $\rho_1 = 1.0, \ \rho_{10} = 0.8$  & 40.41 & 11.134  & 30.17 & 9.582  \\
& $\rho_1 = 1.0, \ \rho_{10} = 0.9$  & 40.36 & 11.310  & 30.15 & 10.023 \\
%\end{tabular}
%}
%\caption{Complete results of {\bf ensemble top-3} method.}%, backing the curves in Figure~\ref{fig:method}}
%\label{tab:esb_top3}
%\end{table}
%\vspace{-15pt}
%
%
%% Done
%\begin{table}[h]
%\resizebox{0.9\linewidth}{!}{%
%\centering
%\begin{tabular} {c | c c | c c}
%\multirow{2}{*}{Hyper-parameters} & \multicolumn{2}{c | }{ZH$\to$EN} & \multicolumn{2}{c}{DE$\to$EN} \\
%& BLEU  & AL & BLEU & AL \\
\hline
\multirow{18}{*}{\rot{ensemble all}}& $\rho_1 = 0.2, \ \rho_{10} = 0.0$  & 26.81 & 1.231   & 24.55 & 2.383 \\
& $\rho_1 = 0.3, \ \rho_{10} = 0.0$  & 32.61 & 3.536   & 25.74 & 2.851 \\
& $\rho_1 = 0.4, \ \rho_{10} = 0.0$  & 35.96 & 5.219   & 26.46 & 3.367 \\
& $\rho_1 = 0.5, \ \rho_{10} = 0.0$  & 37.31 & 6.270   & 26.97 & 3.973 \\
& $\rho_1 = 0.6, \ \rho_{10} = 0.0$  & 38.40 & 6.959   & 27.20 & 4.666 \\
& $\rho_1 = 0.7, \ \rho_{10} = 0.0$  & 38.64 & 7.590   & 27.63 & 5.241 \\
& $\rho_1 = 0.8, \ \rho_{10} = 0.0$  & 39.10 & 8.134   & 27.78 & 5.828 \\
& $\rho_1 = 0.9, \ \rho_{10} = 0.0$  & 39.18 & 8.523   & 27.89 & 6.290 \\
& $\rho_1 = 1.0, \ \rho_{10} = 0.0$  & 38.80 & 8.761   & 27.89 & 6.650 \\
& $\rho_1 = 1.0, \ \rho_{10} = 0.1$  & 38.67 & 9.264   & 27.94 & 7.151 \\
& $\rho_1 = 1.0, \ \rho_{10} = 0.2$  & 38.62 & 9.682   & 27.86 & 7.594 \\
& $\rho_1 = 1.0, \ \rho_{10} = 0.3$  & 38.62 & 10.029  & 27.98 & 8.014 \\
& $\rho_1 = 1.0, \ \rho_{10} = 0.4$  & 38.62 & 10.274  & 28.17 & 8.395 \\
& $\rho_1 = 1.0, \ \rho_{10} = 0.5$  & 38.57 & 10.477  & 28.17 & 8.710 \\
& $\rho_1 = 1.0, \ \rho_{10} = 0.6$  & 38.60 & 10.632  & 28.23 & 8.989 \\
& $\rho_1 = 1.0, \ \rho_{10} = 0.7$  & 38.59 & 10.770  & 28.31 & 9.253 \\
& $\rho_1 = 1.0, \ \rho_{10} = 0.8$  & 38.58 & 10.890  & 28.32 & 9.517 \\
& $\rho_1 = 1.0, \ \rho_{10} = 0.9$  & 38.56 & 11.029  & 28.34 & 9.830 \\
\end{tabular}
}
\caption{Complete results of {\bf wait-if-diff}, {\bf ensemble top-3} and {\bf ensemble all}.}%, backing the curves in Figure~\ref{fig:method}}
\label{tab:esb_all}
\end{table}
\vspace{-15pt}

% Done
\begin{table}[h]
\resizebox{0.99\linewidth}{!}{%
\centering
\begin{tabular} {c@{\ }c | c c | c c}
& \multirow{2}{*}{Hyper-parameters} & \multicolumn{2}{c | }{ZH$\to$EN} & \multicolumn{2}{c}{DE$\to$EN} \\
& & BLEU  & AL & BLEU & AL \\
\hline
\multirow{8}{*}{\rot{wait-if-worse}} & $s_0 = 1, \ \delta = 1$  & 31.67 & 6.857   & 21.77 & 4.930 \\
& $s_0 = 2, \ \delta = 1$  & 32.28 & 7.170   & 22.26 & 5.005 \\
& $s_0 = 4, \ \delta = 1$  & 33.36 & 7.964   & 23.30 & 5.697 \\
& $s_0 = 6, \ \delta = 1$  & 34.78 & 9.319   & 24.27 & 6.914 \\
& $s_0 = 1, \ \delta = 2$  & 36.28 & 12.731  & 26.52 & 10.268 \\
& $s_0 = 2, \ \delta = 2$  & 36.62 & 13.133  & 26.39 & 10.138 \\
& $s_0 = 4, \ \delta = 2$  & 36.89 & 13.629  & 26.68 & 10.806 \\
& $s_0 = 6, \ \delta = 2$  & 37.50 & 14.662  & 27.09 & 11.877 \\
%\end{tabular}
%}
%\vspace{-5pt}
%\caption{Complete results of {\bf wait-if-worse} method.}%, backing the crosses in Figure~\ref{fig:method}}
%\label{tab:wiw}
%\end{table}
%\vspace{-15pt} 
%
%
%
%
%
%% Done
%\begin{table}[h]
%\resizebox{0.9\linewidth}{!}{%
%\centering
%\begin{tabular} {c | c c | c c}
%\multirow{2}{*}{Hyper-parameters} & \multicolumn{2}{c | }{ZH$\to$EN} & \multicolumn{2}{c}{DE$\to$EN} \\
%& BLEU  & AL & BLEU & AL \\
\hline
\multirow{18}{*}{\rot{single}}& $\rho_1 = 0.2, \ \rho_{10} = 0.0$  & 31.24 & 3.335   & 22.72 & 1.989 \\
& $\rho_1 = 0.3, \ \rho_{10} = 0.0$  & 32.96 & 3.781   & 23.85 & 2.211 \\
& $\rho_1 = 0.4, \ \rho_{10} = 0.0$  & 34.39 & 4.455   & 25.05 & 2.672 \\
& $\rho_1 = 0.5, \ \rho_{10} = 0.0$  & 36.23 & 5.254   & 25.61 & 3.047 \\
& $\rho_1 = 0.6, \ \rho_{10} = 0.0$  & 36.75 & 5.750   & 26.73 & 3.627 \\
& $\rho_1 = 0.7, \ \rho_{10} = 0.0$  & 36.95 & 6.526   & 27.21 & 4.187 \\
& $\rho_1 = 0.8, \ \rho_{10} = 0.0$  & 37.67 & 7.030   & 27.84 & 4.785 \\
& $\rho_1 = 0.9, \ \rho_{10} = 0.0$  & 38.41 & 7.604   & 28.41 & 5.330 \\
& $\rho_1 = 1.0, \ \rho_{10} = 0.0$  & 37.89 & 8.021   & 28.81 & 5.813 \\
& $\rho_1 = 1.0, \ \rho_{10} = 0.1$  & 38.45 & 8.458   & 29.02 & 6.169 \\
& $\rho_1 = 1.0, \ \rho_{10} = 0.2$  & 38.20 & 8.839   & 29.20 & 6.596 \\
& $\rho_1 = 1.0, \ \rho_{10} = 0.3$  & 38.59 & 9.386   & 29.32 & 7.042 \\
& $\rho_1 = 1.0, \ \rho_{10} = 0.4$  & 38.81 & 9.805   & 29.19 & 7.581 \\
& $\rho_1 = 1.0, \ \rho_{10} = 0.5$  & 38.77 & 10.141  & 29.29 & 8.079 \\
& $\rho_1 = 1.0, \ \rho_{10} = 0.6$  & 38.75 & 10.463  & 29.21 & 8.589 \\
& $\rho_1 = 1.0, \ \rho_{10} = 0.7$  & 38.76 & 10.733  & 29.25 & 9.044 \\
& $\rho_1 = 1.0, \ \rho_{10} = 0.8$  & 38.51 & 10.944  & 29.19 & 9.491 \\
& $\rho_1 = 1.0, \ \rho_{10} = 0.9$  & 38.49 & 11.201  & 29.10 & 9.972 \\
%\end{tabular}
%}
%\vspace{-5pt}
%\caption{Complete results of {\bf single} method.}%, backing the curves in Figure~\ref{fig:method}}
%\label{tab:single}
%\end{table}
%\vspace{-15pt} 
%% Done
%\begin{table}[h]
%\resizebox{0.85\linewidth}{!}{%
%\centering
%\begin{tabular} {c | c c | c c}
%\multirow{2}{*}{Hyper-parameters} & \multicolumn{2}{c | }{ZH$\to$EN} & \multicolumn{2}{c}{DE$\to$EN} \\
%& BLEU  & AL & BLEU & AL \\
\hline
\multirow{10}{*}{\rot{wait-$k$}}& $k = 1$  & 28.30 & 2.968   & 21.31 & 1.695  \\
& $k = 2$  & 30.74 & 3.519   & 23.10 & 2.652  \\
& $k = 3$  & 32.45 & 5.076   & 25.22 & 3.768  \\
& $k = 4$  & 33.80 & 5.896   & 26.29 & 4.697  \\
& $k = 5$  & 34.67 & 7.041   & 27.42 & 5.771  \\
& $k = 6$  & 35.80 & 8.175   & 27.73 & 6.658  \\
& $k = 7$  & 36.77 & 9.033   & 28.53 & 7.569  \\
& $k = 8$  & 37.49 & 9.542   & 28.64 & 8.548  \\
& $k = 9$  & 38.17 & 10.560  & 28.92 & 9.379  \\
& $k = 10$ & 38.44 & 11.337  & 29.06 & 10.261 \\
%\end{tabular}
%}
%\vspace{-5pt}
%\caption{Complete results of {\bf wait-$k$} method.}%, backing the curves in Figure~\ref{fig:method}}
%\label{tab:waitk}
%\end{table}
%\vspace{-15pt} 
%
%% Done
%\begin{table}[h]
%\resizebox{0.85\linewidth}{!}{%
%\centering
%\begin{tabular} {c | c c | c c}
%\multirow{2}{*}{Hyper-parameters} & \multicolumn{2}{c | }{ZH$\to$EN} & \multicolumn{2}{c}{DE$\to$EN} \\
%& BLEU  & AL & BLEU & AL \\
\hline
\multirow{10}{*}{\rot{test-time wait-$k$}}& $k = 1$  & 27.54 & 2.884   & 21.84 & 3.204  \\
& $k = 2$  & 29.57 & 3.873   & 22.64 & 3.954  \\
& $k = 3$  & 30.70 & 5.103   & 22.96 & 4.729  \\
& $k = 4$  & 31.37 & 5.941   & 23.60 & 5.558  \\
& $k = 5$  & 32.67 & 6.993   & 24.48 & 6.412  \\
& $k = 6$  & 33.92 & 8.051   & 24.92 & 7.298  \\
& $k = 7$  & 34.16 & 8.850   & 25.23 & 8.144  \\
& $k = 8$  & 34.95 & 9.720   & 25.48 & 9.025  \\
& $k = 9$  & 35.34 & 10.566  & 26.05 & 9.867  \\
& $k = 10$ & 35.87 & 11.383  & 26.28 & 10.699 \\
\end{tabular}
}
%\vspace{-5pt}
\caption{Complete results of {\bf wait-if-worse}, {\bf single}, {\bf wait-$k$} and {\bf test-time wait-$k$}.}%, backing the curves in Figure~\ref{fig:method}}
\label{tab:test_waitk}
\end{table}
\vspace{-15pt}

\end{document}